\documentclass[letterpaper, 10 pt, conference]{ieeeconf}  

\IEEEoverridecommandlockouts                              

\usepackage{graphics} 
\usepackage{epsfig} 
\usepackage{xcolor}
\usepackage{amsmath} 
\usepackage{amssymb}  
\usepackage{flushend}   

\newtheorem{theorem}{Theorem}[section]

\newtheorem{definition}{Definition}[section]
\newtheorem{lemma}[theorem]{Lemma}

\usepackage{fancyhdr}
\title{\LARGE \bf
An ensemble of online estimation methods for one degree-of-freedom models of unmanned surface vehicles: applied theory and preliminary field results with eight vehicles 
}

\author{Tyler M. Paine$^{1,2}$ and Michael R. Benjamin$^{1}$
\thanks{$^{1}$Department of Mechanical Engineering, Massachusetts Institute of Technology, 
        Cambridge, MA 02139, USA
        {\tt\small tpaine@mit.edu, mikerb@mit.edu}}%
\thanks{$^{2}$Woods Hole Oceanographic Institution
        Woods Hole, MA 02543, USA}%
}

\begin{document}

\maketitle
\thispagestyle{empty}
\pagestyle{empty}

\begin{abstract}
In this paper we report an experimental evaluation of three popular methods for online system identification of unmanned surface vehicles (USVs) which were implemented as an ensemble: certifiably stable shallow recurrent neural network (RNN), adaptive identification (AID), and recursive least squares (RLS).  The algorithms were deployed on eight USVs for a total of 30 hours of online estimation.  During online training the loss function for the RNN was augmented to include a cost for violating a sufficient condition for the RNN to be stable in the sense of contraction stability.   Additionally we described an efficient method to calculate the equilibrium points of the RNN and classify the associated stability properties about these points.  We found the AID method had lowest mean absolute error in the online prediction setting, but a weighted ensemble had lower error in offline processing.  
\end{abstract}

\thispagestyle{fancy}
\section{INTRODUCTION}
Accurate dynamical models of marine robotic vehicles are needed for many modern approaches to control, navigation, and state estimation. 
The process of system identification is often performed in an offline setting, or after experimental data has been collected and filtered.  
In contrast, online model identification or parameter estimation is more difficult because sensor data occasionally contain outliers which are difficult to recognize in real-time, sensor data can also be asynchronous, robotic vehicles have limited computational resources, and there are more severe consequences if performance is unstable.  
Although it is more difficult, online learning allows for more sophisticated autonomous behaviors, such as real-time fault recovery or dynamic re-weighting of objectives. When accurate, online estimation can be much more affordable than traditional methods that require tow-tanks or instrumented ranges.  This later point is increasingly more important as the number and variety of unmanned marine vehicles increases.

This paper reports the performance of an ensemble of online methods for system identification which were implemented on a fleet of autonomous vehicles.  The basic concept is to combine in parallel three common approaches to system identification: recurrent neural network (RNN), adaptive identification (AID), and a recursive least squares (RLS) thruster map, into a single ensemble estimate of the dynamical model of a small unmanned surface vehicle (USV).  

\begin{figure}[h]
  \centering
  \includegraphics[width=1\columnwidth]{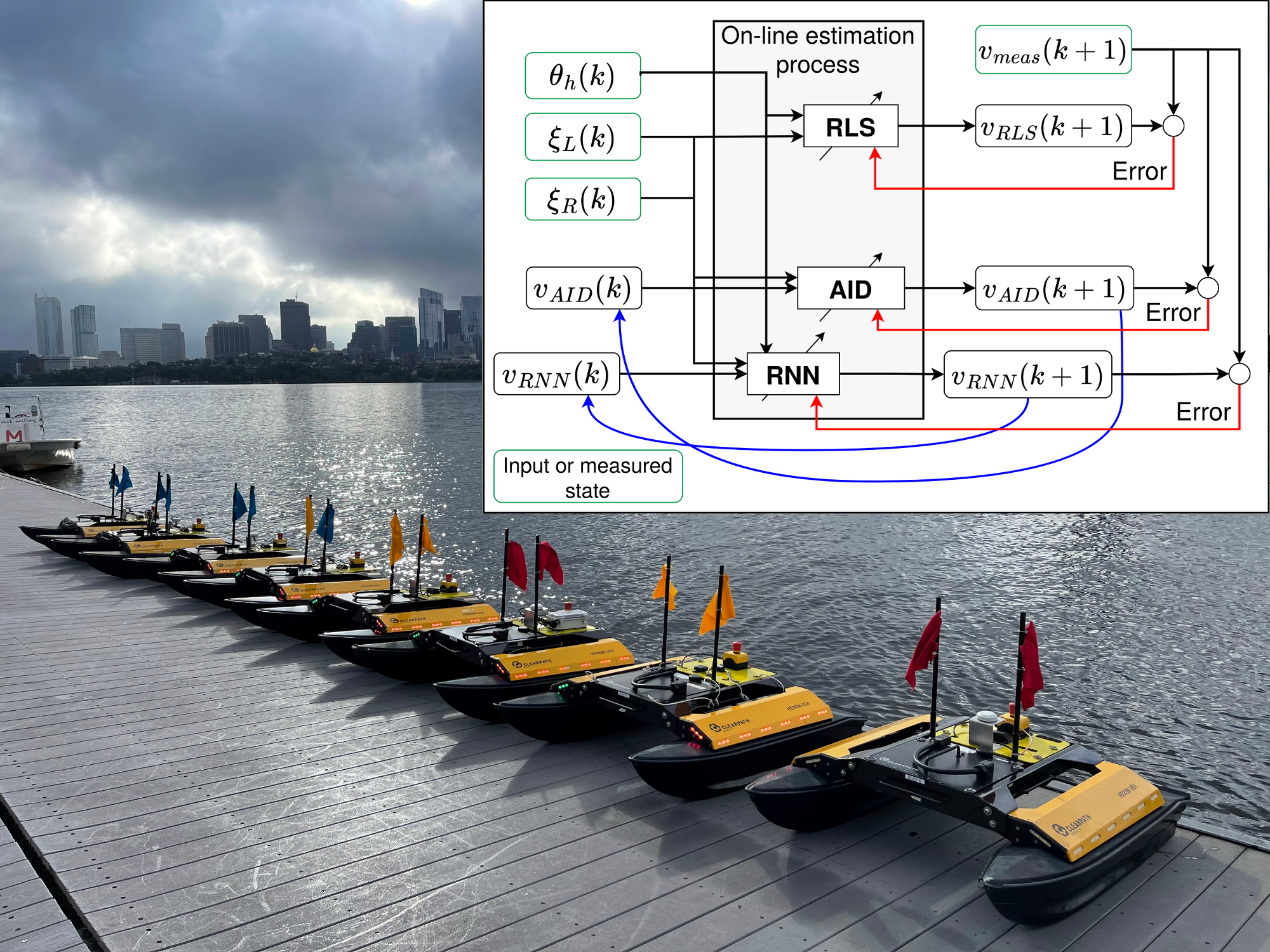}
  \caption{Eight Heron Unmanned Surface Vehicles (USVs) made by Clearpath Robotics were used for field testing. Inset: Block diagram of online ensemble estimation using adaptive identifier (AID), recursive least squares (RLS), and recurrent neural network (RNN) approaches. }
  \label{fig:HeronRobots}
  \vspace{-2mm}
\end{figure}

The main contributions of this work are the following:
\begin{itemize}
\item Report and compare experimental performance of an ensemble of RLS, AID, and RNN approaches to online system identification for one degree-of-freedom (1-DOF) models across a fleet of eight small USVs.
\item Show a 1-DOF model of USV motion can be modeled as a contracting system, and use a modified certification criteria as a sufficient condition for the shallow RNN to be state-stable during online estimation.
\end{itemize}
In this preliminary study we fielded algorithms that have well understood stability properties.  Safety is important because our autonomous vehicles routinely operate on the busy public Charles River, and the ensemble algorithms must run ``in the background'' on microprocessors which are simultaneously computing other tasks, e.g behavior optimization or perception. 

We deployed the system identification algorithms onboard a fleet of eight Heron USVs made by Clearpath Robotics shown in Figure \ref{fig:HeronRobots}. The dynamics of each of the Heron vehicles are slightly different, likely due to over seven years of extensive use for both coursework and research.  Some hulls slowly leak, and some thrusters and motor controllers are more worn than others.  For these reasons, the Heron USVs are excellent, low risk candidates to test the three online estimation approaches.

\section{RELATED WORK}

In this section we review recent publications that describe different methods of model identification, and highlight were online learning was reported using real unmanned marine vehicles.  Several online implementations of RLS estimation of dynamical parameters have been reported for underwater vehicles \cite{RandeniP2018NonlinearDynamics} and aerial vehicles \cite{birnbaum2014UAS}. In the case of Randeni et al. in \cite{RandeniP2018NonlinearDynamics}, the RLS estimation is part of a navigation estimation process, whereas the work of Birnbaum et al. in \cite{birnbaum2014UAS} is primarily focused on fault detection.  Least-squares estimation for unmanned underwater vehicles (UUV)s is commonly used in offline, or in a post-processing scenario; a few relevant examples can be found in \cite{ijoe2014martin, 2007hegrenaes, graver2003underwater}.  The theory for stable AID of underwater vehicles has been established, which has largely been validated by experiments with UUVs in an offline setting. AID approaches for 1-DOF \cite{cst2003smallwood}, 3-DOF \cite{paine2018Oceans}, and 6-DOF \cite{mcfarlandJAC2021, harrismaopaine2022IJRR} models of UUV motion have been reported. 

Most classic machine learning techniques do not use physical models.  These include support vector regression (SVR), Gaussian process regression (GPR), and neural networks (NN).  Wehbe et al. in \cite{Wehbe2017icra} report a comparison of the performance of these three methods to predict dynamics of an UUV in an offline study.  However, the same authors in \cite{Wehbe2019icra} develop a framework for online estimation, and an incremental SVR strategy is implemented on-board the UUV Dragon. There are many examples where offline machine learning is used for system identification for UUVs including \cite{sonnenburg2013JFR, yuh1999ICRA}, and an interesting hybrid approach combining least squares with long short term memory (LSTM) networks for USVs is reported in \cite{woo2018AOR}

\section{MATHEMATICAL NOTATIONS AND DEFINITIONS}
\begin{definition} \label{def:recurrmodel}\cite{miller2018stable}
\emph{A recurrent model is a non-linear dynamical system given by a differentiable state-transition map $\phi_w(): {\rm I\!R}^q \times {\rm I\!R}^m \rightarrow {\rm I\!R}^q$,  parameterized by $\vec{w} \in {\rm I\!R}^{\bar{n}}$. The state $\vec{x} \in  {\rm I\!R}^q$ evolves in discrete time steps according to the update rule 
\begin{align}
\vec{x}_{k+1} =& \phi_w(\vec{x}_k, \vec{u}_{k}),
\end{align}
where the vector $u_{k+1} \in {\rm I\!R}^m$ is the system input vector.  }
\end{definition}

\begin{definition} \label{def:contracting}
\cite{lohmiller1998contraction} \emph{A recurrent model (\ref{def:recurrmodel}) is $K-$contracting on $x$ if there exists some $K < 1$ such that, for any weights 
$\vec{w} \in {\rm I\!R}^{\bar{n}}$, states $\vec{x}_1$, $\vec{x}_2 \in  {\rm I\!R}^q$, and input $\vec{u} \in {\rm I\!R}^m$,
\begin{align}
||\phi_w(\vec{x}_1, \vec{u}) - \phi_w(\vec{x}_2, \vec{u})|| \leq& K || \vec{x}_1 - \vec{x}_2||.
\end{align}
}
\end{definition}

\section{ONLINE SYSTEM IDENTIFICATION}
\subsection{1-DOF model of USV motion}
For the AID approach we used the commonly accepted model of USV motion in the surge degree-of-freedom \cite{fossen_1994}:
\begin{align}
m \dot{v}(t) =& c_{q} | v(t) | v(t) + c_{l} v(t) + c_{\dot{\theta}} |\dot{\theta}(t) v(t)|\\
&+ \tau_{L}(\xi_L(t), v(t)) + \tau_{R}(\xi_R(t), v(t))\label{eq:simple1dof}
\end{align}
where the system state is the velocity in the surge degree of freedom is $v(t)\in {\rm I\!R}$ and parameters for total mass $m \in {\rm I\!R}_{+}$, quadratic drag $c_q \in {\rm I\!R}_{-}$, linear drag $c_l \in {\rm I\!R}_{-}$, and cross-term drag $c_{\dot{\theta}} \in {\rm I\!R}_{-}$. The mass term $m$ includes the rigid-body and added mass terms.  The heading of the vehicle is $\theta$. The parameters for the thrust generated by each of the two thrusters is parameterized as 
\begin{align}
    \tau_i(\xi_i(t), v(t)) =& \alpha_i \xi_i(t) + \beta_i \xi_i^2(t) \\
    &+ \gamma_i \xi_i(t) v(t) \ \text{for} \ i = L, R,
\end{align} 
where parameters $\alpha_i, \beta_i \in {\rm I\!R}_{+}, \  \gamma_i \in {\rm I\!R}_{-} $ and $\xi_i(t)$ is the \emph{commanded} thruster rotational speed (since unfortunately the propeller shaft was not directly instrumented, and therefore the actual speed is not precisely known).  The notation $L,R$ indicates values for the left or right thruster respectively. Our model of the thrust function $\tau$ included a term that is quadratic in commanded propeller speed, which is the typical model of a propeller operating close to a ``Bollard pull'' condition.  We included a linear term, since the true propeller speed is not precisely known.  Finally, we included a third term to capture the theoretically inverse relationship between of forward speed and the thrust force.  The thrusters on the Heron USVs have ducted inlets, and the water mass entering the inlet is a function of forward speed.  

When implemented, the algorithms operated in discrete time, and with a time step $dt$ the discrete approximation of (\ref{eq:simple1dof}) can be expressed as
\begin{align}
v_{k+1} =& f_D\big( v_k, \xi_{L_{k+1}}, \xi_{R_{k+1}} \big)
\end{align}
Hereafter we drop the explicit dependence of signals on time for clarity. 

\subsection{RNN system identification}
Consider the following commonly used shallow RNN structure with $n$ artificial neurons for a single recursive state $x\in {\rm I\!R}$, and inputs $\vec{u} = \begin{vmatrix} u_1 & u_2 & \hdots u_n \end{vmatrix} \in {\rm I\!R}^{m}$: 
\begin{align}
x_{k+1} =& \vec{a}^T \sigma \big(\vec{w}_1 x_{k} + W_2 u_{k+1} + \vec{b} \big), \label{eq:RNNDef}
\end{align}
where the matrix $W = \begin{bmatrix} \vec{w}_1 & W_2 \end{bmatrix} \in {\rm I\!R}^{n \times (m+1)}$ contains the connection weights for the fully connected single layer, and $\vec{b} \in {\rm I\!R}^{n}$ contain the bias for each artificial neuron.  The activation function $\sigma(): {\rm I\!R}^{n} \rightarrow {\rm I\!R}^{n}$ is the rectified linear unit (RELU). 
Finally $\vec{a} \in {\rm I\!R}^{n}$ is a vector of connection weights from the single layer to the output. 

The RNN structure (\ref{eq:RNNDef}) used in this study was not especially complex.  However, in the following sections we will
\begin{enumerate}
    \item Use the simplicity to clearly define stability properties.  Once experimentally verified, these approaches to verifying stability may be extended to more complex RNN architectures. 
    \item Show that the RNN structure (\ref{eq:RNNDef}) is already more complex in terms of number of parameters and computational time than the alternative AID and RLS methods described in Sections \ref{sec:AID} and \ref{sec:RLS} respectively. 
\end{enumerate}

To address the first item, we will apply the results of recent research on the topic of contraction analysis of RNNs \cite{TSUKAMOTO2021135}, and we call an RNN which has the contraction property a ``contracting RNN''.

\subsubsection{Contracting model of USV motion} \label{sec:cont_USV_motion}
First, to justify the use of a contracting RNN to identify the system dynamics of a USV we must show that the system can be reasonably represented by a contracting model per Definition \ref{def:contracting}. In this situation the ``true'' system dynamics are unknown, and the model (\ref{eq:simple1dof}) is a limited - but accepted - representation of the actual high-dimensional system.  However, for this study we use the model (\ref{eq:simple1dof}) to show it is reasonable to represent the system dynamics as a discrete contracting model.

We restrict velocities in the surge direction to be non-negative, which is a typical operating range for the Heron USVs.  Without loss of generality, consider the linear state transformation $x = h \cdot v$, where $h = \frac{1}{2 v_{max}} \in {\rm I\!R}_+$, that scales the domain of surge velocity $v \in [0, v_{max}]$ to the range $x \in [0, 0.5]$.  Consider any transformed states $x_1 = h \cdot v_1$, $x_2 = h \cdot v_2$ of the USV motion model (\ref{eq:simple1dof}), then for a discrete time step $dt$

\begin{align}
\Big| f_D\big(x_1, & \xi_L, \xi_R \big)  - f_D\big(x_2, \xi_L, \xi_R \big) \Big| \\
=&
\Big| \frac{2 v_{max} c_q dt}{m} \big( x_1^2 - x_2^2 \big) \\
&+ \bigg( \frac{ (c_l + c_{\dot{\theta}} + \gamma_R \xi_R + \gamma_L \xi_L ) dt}{m} + 1 \bigg) \big( x_1 - x_2 \big) \Big| \\
=& \Big| \bar{a} \big( x_1 - x_2 \big) \Big|, 
\end{align}
where
\begin{align}
\bar{a} =& \frac{dt}{m} \Big( ( 2 v_{max} c_q ) \frac{ x_1^2 - x_2^2}{x_1 - x_2}  \\
&+  (c_l + c_{\dot{\theta}} + \gamma_R \xi_R + \gamma_L \xi_L ) \Big) + 1.
\end{align}
If we can show $-1 < \bar{a} < 1$ for a reasonable choice of $dt>0$, then the system is contracting by Definition \ref{def:contracting}.  The term $ \frac{ x_1^2 - x_2^2}{x_1 - x_2} = (x_1 + x_2) \in [0,1] \ \forall x_1, x_2 \in [0,0.5]$.  We note that parameters $c_q, c_l, c_{\dot{\theta}}, \gamma_R, \gamma_L \in {\rm I\!R}_{-}$, so any choice of $dt>0$ would satisfy the upper bound of $\bar{a} <1$ for any inputs $\xi_{R}, \xi_L$. We are left to satisfy the lower bound which can be written as

\begin{align}
-2 &< \frac{dt}{m} \Big( ( 2 v_{max} c_q ) \frac{ x_1^2 - x_2^2}{x_1 - x_2} +  (c_l + c_{\dot{\theta}} + \gamma_R \xi_R + \gamma_L \xi_L ) \Big) \\
 \frac{dt}{m}  &< \frac{2}{\Big( ( 2 v_{max} c_q ) \frac{ x_1^2 - x_2^2}{x_1 - x_2} +  (c_l + c_{\dot{\theta}} + \gamma_R \xi_R + \gamma_L \xi_L ) \Big)}. \label{eq:ineqbound}
\end{align}
Choosing 
\begin{align}
dt < \frac{2m}{\Big|2 v_{max} c_q  +  c_l + c_{\dot{\theta}} + (\gamma_R + \gamma_L) \xi_{max}  \Big|}
\end{align}
satisfies (\ref{eq:ineqbound}) where $\xi_{max} \in {\rm I\!R}_{+} $ is the upper bound of the control input, e.g. $\xi_{max} = \text{max}\big(\xi_R(t), \xi_L(t)\big) \ \forall t$. 

These restrictions are reasonable given previous studies of marine vessel dynamics \cite{2007hegrenaes,martinthesis}. We confirmed this assumption is reasonable for the Heron USVs by performing preliminary parameter estimation in an offline setting during the 2021 summer season and found the restriction to be $0<dt<\approx 2$ seconds.

\subsubsection{Stability of RNNs}
The goal of this section is to define a property of a shallow RNN (\ref{eq:RNNDef}) that is a sufficient condition for the the RNN to be stable in the sense of contraction stability.  
In addition, we will define the stability properties of the transformed state variable $x$ about equilibrium points, and describe an efficient methodology to estimate these points in a trained RNN. These are two separate tasks.  

\begin{theorem} \label{th:RNNcont}
The RNN model (\ref{eq:RNNDef}) is contracting with the contraction metric $p \in {\rm I\!R}_+$ if there exists a diagonal matrix $\Lambda \geq 0$ with diagonal entries $\lambda_i \in {\rm I\!R}_+$ such that 
\begin{align}
M = \begin{bmatrix}
2 - p & 0 & -a^T \\
0 & p & - \vec{w}^T_1 \Lambda \\
-a & -\Lambda \vec{w}_1 & \Lambda
\end{bmatrix} \geq 0 \label{eq:LMI_const}
\end{align}
\emph{Proof: See \cite{manchester2021CiRNNTutorial, Revay2020ContractingRNN, TSUKAMOTO2021135}}
\end{theorem}

The statement (\ref{eq:LMI_const}) is usually implemented as an LMI constraint satisfaction problem, or as a semi-definite programming problem \cite{Revay2020ContractingRNN}.  

Instead, we propose an simpler alternate sufficient condition for contraction that uses the Levy-Desplanques Theorem, which is in the spirit of previous work by Jin and Gupta \cite{jin1999stable}.  
This alternate condition is given as Lemma \ref{lem:theoremDisk}.

\begin{lemma} \label{lem:theoremDisk}
The RNN model (\ref{eq:RNNDef}) is contracting with the contraction metric $p \in (0,2)$ if there exists $\lambda_i$ as defined in Theorem \ref{th:RNNcont} such that
\begin{align}
2-p >& \sum_{i}^n |a_i| \label{eq:diag_const1} \\
p   >& \sum_{i}^n  \lambda_i |w_{1i} | \label{eq:diag_const2} \\
\lambda_i >& |a_i| + \lambda_i |w_{1i} | \label{eq:diag_const3} 
\end{align}
\emph{Proof: From Ger\v{s}gorin's theorem, the eigenvalues of $M$ are in the union of the Ger\v{s}gorin discs, which are bounded away from zero by constraints (\ref{eq:diag_const1}-\ref{eq:diag_const3}) and $p \in (0,2)$.  The matrix $M$ is symmetric real and $M_{ii} > 0 \ \forall i = 1, \hdots, n$, and therefore $M$ is positive definite. (\cite{horn2012matrix}, Theorem 6.1.10)}
\end{lemma}

Here we present our a simplified - but more conservative - sufficient condition for the contraction property of the RNN model (\ref{eq:RNNDef}).  
\begin{theorem} \label{th:RNNSimple}
There exists $\lambda_i = \frac{1}{n+1}, \forall i = 1,2, \hdots n$ as defined in Theorem \ref{th:RNNcont} such that the RNN model (\ref{eq:RNNDef}) with $n>1$ artificial neurons is contracting with the metric $p = 1$ if
\begin{align}
|| |\vec{a}| + (\frac{1}{n+1}) |\vec{w}_1| ||_{\infty} < \frac{1}{n+1} \label{eq:simpleConst}
\end{align}
\emph{Proof: It suffices to show that the constraint (\ref{eq:simpleConst}) meets the constraints (\ref{eq:diag_const1}-\ref{eq:diag_const3}) in Lemma \ref{lem:theoremDisk} which are are sufficient conditions for the contraction property.  By inspection (\ref{eq:simpleConst}) satisfies (\ref{eq:diag_const3}). From (\ref{eq:simpleConst}) we can say that for $|a_i| \geq 0 \forall i$ then $|| \vec{w}_1||_{\infty} \leq 1$, and $|| \vec{w}_{1} ||_1 \leq n|| \vec{w}_{1}||_{\infty} \leq n$.  Thus we can show (\ref{eq:simpleConst}) satisfies (\ref{eq:diag_const2}) since
\begin{align}
\sum_{i}^n  \lambda_i |w_{1i} | =  \lambda || \vec{w}_{1} ||_1 \leq n \lambda || \vec{w}_{1}||_{\infty} \leq n \lambda <& \frac{n}{n+1} < 1 = p  \label{eq:step1}
\end{align}  
In a similar manner, we can use (\ref{eq:simpleConst}) to bound $||\vec{a}||_1$, since for $|\vec{w}_{1i}| \geq 0$ then $||\vec{a}||_1 \leq n \lambda$.  Thus we can show (\ref{eq:simpleConst}) satisfies (\ref{eq:diag_const1}) since 
\begin{align}
\sum_{i}^n |a_i| = ||\vec{a}||_1 \leq n \lambda <& \frac{n}{n+1} < 1 = 2-p
\end{align}}
\end{theorem}

Remark: The equality assumption $\lambda_i = \lambda \quad \forall i = 1,2, \hdots n$ essentially enforces equal bounds on every weight $\vec{a}_i$ and $\vec{w}_{1i}$ in the RNN.  This fact is why the conditions in Theorem \ref{th:RNNSimple} are more conservative.

The sufficient conditions defined in Theorem \ref{th:RNNSimple} were added as additional costs for the ADAM weight optimizer, similar to the approaches reported in \cite{jin1999stable,Revay2020ContractingRNN}.  Specifically, the cost is defined as
\begin{align}
\mathcal{L} = \frac{1}{2}(x_{k+1} - x_{meas})^2 + \frac{\eta}{2} \sum_{i=1}^n \psi_i (a_i^2 + w_{1i}^2),
\end{align}
where the constraint enforcing multiplier 
\begin{align}
\psi_i = 
\begin{cases}
1 & \text{if} \ |a_i| + \frac{1}{n+1} |w_{1i}| > \frac{1}{n+1} \\
0 & \text{otherwise}
\end{cases}
\end{align}
and $\eta > 0$ is the learning weight associated with the contraction property. 

\subsubsection{Equilibrium points of a trained shallow RNN} \label{sec:equalib_points_RNN}
Here we restate results in theory of discrete systems from \cite{galor2007discrete} and \cite{jin1999stable} and apply them to the analysis of stability of a contracting RNN about equilibrium points.
\begin{definition} \label{def:equalib poins}
\emph{If we consider the trained RNN (\ref{eq:RNNDef}) to be a recurrent model (\ref{def:recurrmodel})
then the equilibrium points, $x_{eq1}, x_{eq2}, \hdots x_{eq(n+1)}$, can be defined as 
\begin{align}
x_{eqi} = \phi_w \big( x_{eqi},  \vec{0} \big) \quad \forall i = 1,2,\hdots,(n+1).  \label{eq:RNNequalib}
\end{align}}
\end{definition}
We will later show that the system (\ref{eq:RNNDef}) can have at most $n+1$ equilibrium points. 

\begin{figure}[ht]
	\centering
  \includegraphics[width=0.9\columnwidth]{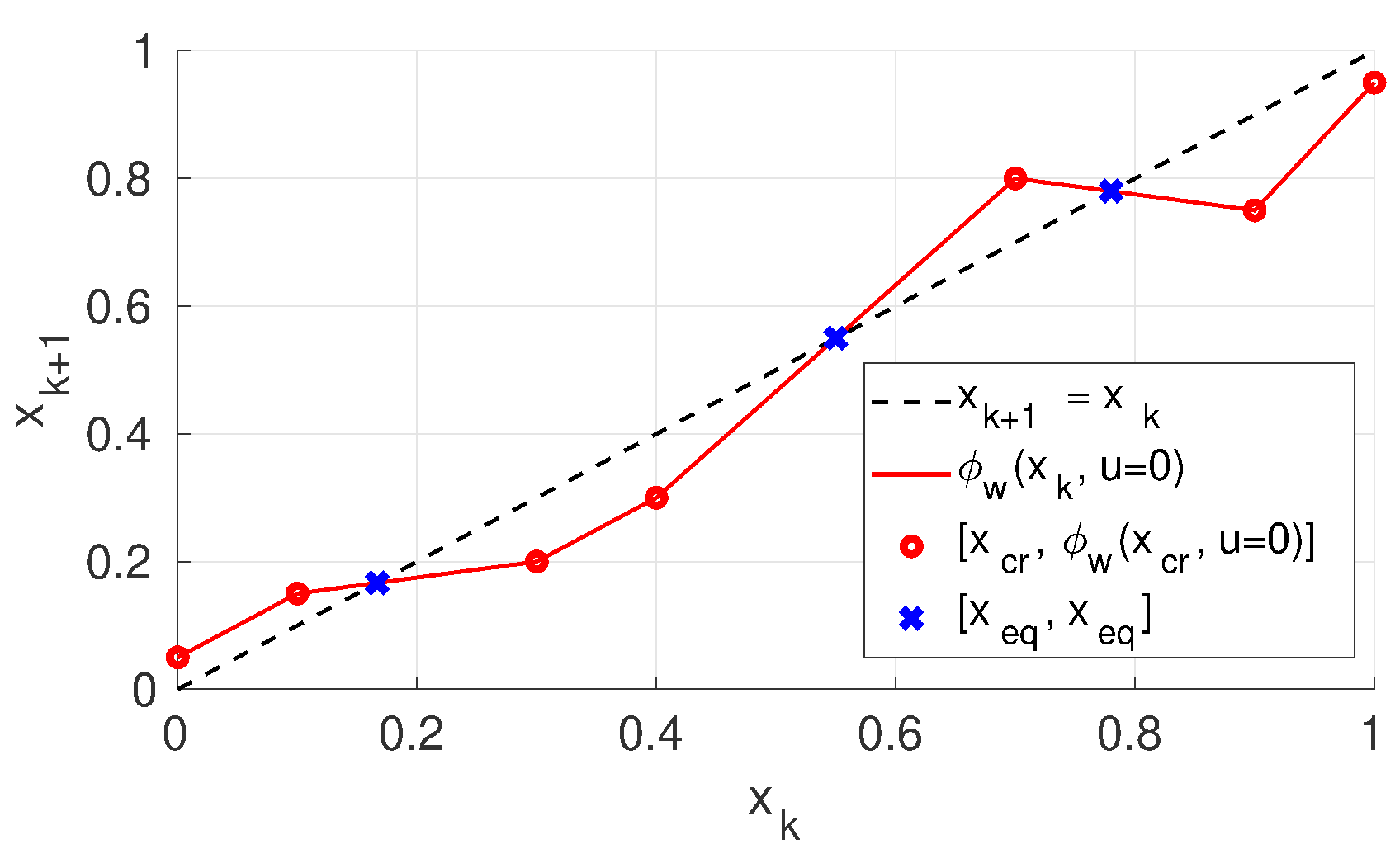}
  \caption{Illustration of the state response from a hypothetical trained shallow RNN with a single recurrent state.  The critical values (\ref{eq:critical_points}) and equilibrium points are shown to bring clarity to the arguments in Sections \ref{sec:equalib_points_RNN} and \ref{sec:equalib_stability_RNN}.  This RNN clearly does not have the contraction property}
  \label{fig:RNNstab}
  \vspace{-5mm}
\end{figure}

The trained RNN (\ref{eq:RNNDef}) is essentially a piece-wise linear function that, if properly optimized, theoretically approximates the (unknown) target recurrent function.  For a shallow RNN with a single recurrent state we can geometrically find equilibrium points by plotting the the function $x_{k+1} = \phi_w \big( x_k, \xi_R=0, \xi_L=0 \big)$ which by inspection of (\ref{eq:RNNDef}) must be bijective. A hypothetical plotted example is shown in Figure \ref{fig:RNNstab}. 

It is clear from Figure \ref{fig:RNNstab}, the entire function $x_{k+1} = \phi_w \big( x_k, \vec{u}_k = \vec{0} \big)$, can be known from just the critical values of $x$ marked as $x_{cr} \in S$, where 
\begin{align}
S = \bigg\{ x_{cr} \in \rm I\!R \ \big| \ x_{cr} = 0,1,& \dfrac{-b_i}{w_{1i}}, \ \ i \in \mathbb{N} < n, \\
& and  \ 0 \leq x_{cr} \leq 1  \bigg\} \label{eq:critical_points}
\end{align}
Critical values $x_{cr}$ satisfy $0 = w_{1i} \cdot x_{cr} + b_i$ , which are the critical values for the $RELU()$ activation function of the $i^{th}$ neuron.  The input bounds are also included.   
With only $n$ calculations we can recover the entire function $x_{k+1} = \phi_w \big( x_k, \vec{u}_k = \vec{0} \big)$.  We can then geometrically determine the equilibrium points $x_{eqi}$ (\ref{eq:RNNequalib}) by calculating the point of intersection with the line of equilibrium as shown in Figure \ref{fig:RNNstab}.  
Furthermore, since the set $S$ contains at most $n+2$ critical values for $x_k$, there are at most $n+1$ possible equilibrium points.   

\subsubsection{Stability about equilibrium points} \label{sec:equalib_stability_RNN}
Here we summarize well understood notions of stability and apply them to the equilibrium points we compute using the technique described in Section \ref{sec:equalib_points_RNN}.  The stability properties of these equilibrium points are defined by the slope at the equilibrium point \cite{galor2007discrete}.  Consider the slope of the piece-wise linear function (\ref{eq:RNNequalib}) as
\begin{align}
q_i = \frac{d}{dx} \Big( \phi_w \big( x_{ei},  \xi_L=0, \xi_R=0 \big) \Big) \bigg|_{x = x_{ei}}
\end{align} 
where
\begin{definition} \cite{galor2007discrete} \label{def:stability}
\~\
\begin{itemize}
\item $q_i \in [0,1) \rightarrow $ locally stable
\item $q_i \in (-1,0) \rightarrow $  locally stable oscillatory convergence
\end{itemize}
\end{definition}
\begin{lemma} \label{lem:RNNcontStability}
A RNN described by (\ref{eq:RNNDef}) that has the contracting property per Definition \ref{def:contracting} implies that $|q_i| < 1 \ \forall x \in [0,1]$, and therefore is either locally stable, or locally stable with oscillatory convergence on the entire domain $x \in [0,1]$.  \\
\emph{Proof: This lemma follows directly from Definition \ref{def:contracting} and Definition \ref{def:stability}.}
\end{lemma}
We must be clear that Lemma \ref{lem:RNNcontStability} does not imply that a contracting RNN (\ref{eq:RNNDef}) has \emph{any} equilibrium points.  However, a contracting system can have at most one equilibrium point \cite{lohmiller1998contraction}, which leads to the following useful lemma regarding stability of RNNs. 
\begin{lemma}
If an RNN described by (\ref{eq:RNNDef})  has one equilibrium point $x_{ei}$, and has the contracting property per Definition \ref{def:contracting}, then the state $x$ is either locally stable, or locally stable with oscillatory convergence about the equilibrium point $x_{ei}$ on the entire domain $x \in [0,1]$. \\
\emph{Proof: This lemma follows directly from Lemma \ref{lem:RNNcontStability}, and the fact that contracting systems have at most one equilibrium point \cite{lohmiller1998contraction}.}
\end{lemma}

\subsubsection{RNN stability summary}
In this section we found a sufficient condition for the RNN (\ref{eq:RNNDef}) to be stable in the sense of contraction stability, and we incorporated this condition into the loss function during training. 
We also described an efficient method to calculate the equilibrium points of a trained RNN and classify the associated stability properties local to those equilibrium points.  In the literature, this process is often referred to as ``certifying'' the safe performance of an RNN model. 

However, we must be clear that these methods do not guarantee that the learned RNN weights converge to the ``true'' weights.  For instance there is an obvious lack of a sufficient condition akin to the persistent excitation condition in adaptive systems.
We also made no assertion that the RNN was minimally parameterized.  More theoretical work is needed to overcome these deficiencies, and that is beyond the scope of this paper.

\subsection{Adaptive identification of system dynamics} \label{sec:AID}
In this preliminary work, we implemented a 1-DOF AID inspired by Smallwood and Whitcomb \cite{cst2003smallwood}, but where the mass parameter (rigid-body and added mass) is assumed to be known and the thruster parameters are unknown.  The continuous time model (\ref{eq:simple1dof}) can be expressed as
\begin{align}
 m\dot{v} =& \begin{bmatrix}
| v | v & v & |\dot{\theta}v| & \xi_L & \xi_L^2 & \xi_L v & \xi_R & \xi_R^2 & \xi_R v 
\end{bmatrix} \\
 m\dot{v} =& \mathcal{W}( v, \omega_{L}, \omega_{R})^T \vec{\theta}_p,
\end{align}
where
\begin{align}
\vec{\theta}_p = 
\begin{bmatrix}
c_q & c_l & c_{\dot{\theta}} & \alpha_L & \beta_L & \gamma_L & \alpha_R & \beta_R & \gamma_R
\end{bmatrix}^T,
\end{align}
$\vec{\theta}_p \in {\rm I\!R}^{9}$ and $\mathcal{W}( v, \omega_{L}, \omega_{R})$ is the AID parameter regressor matrix.  
The parameter update law and stability analysis of this scalar plant AID can be found in \cite{cst2003smallwood}.  

In the future, we plan to simultaneously estimate all parameters - including the mass parameter $m$ - using the new Nullspace-Based AID method \cite{harrismaopaine2022IJRR}.  In this preliminary work we assumed the mass term $m \approx 8 | c_q|$  which was separately estimated using preliminary free-decay tests, and agrees with the parameters given by the manufacture's own simulation model \cite{ClearpathHeronRepo}.

\subsection{Recursive least squares parameter estimation}\label{sec:RLS}
For RLS estimation we considered a quasi-static approximation for the dynamic model of 1-DOF of vehicle motion (\ref{eq:simple1dof}), where $\frac{d}{dt}v = 0$.  With this assumption 
\begin{align}
v \approx \tilde{v} = f_2 \big(\xi_R, \xi_L, \theta_{h}, \dot{\theta}_{h} \big),
\end{align}
where $f_2$ is an unknown non-linear function that maps thruster inputs, heading $\theta$, and heading rate $\dot{\theta}$ to vehicle velocity.  This solution was similar to a thruster mapping, but it also captured the effect of water current velocity on vehicle motion.  A similar approach was used by Hegren{\oe}s et al. in \cite{2007hegrenaes} and Randeni et al. in \cite{RandeniP2018NonlinearDynamics}.  We can approximate the function $ \tilde{v}$ using the polynomial and trigonometric basis functions up to third order, e.g. 
\begin{align}
\tilde{v} =& \zeta_1 \xi_R + \zeta_2 \xi_R^2 + \zeta_3 \xi_R^3 + \zeta_4 \xi_L + \zeta_5 \xi_L^2 + \zeta_6 \xi_L^3 \\
& + \zeta_7 sin(\theta) + \zeta_8 cos(\theta) + \zeta_9 |\dot{\theta}|  \\
\tilde{v} =& \Phi^T
\begin{bmatrix}
\zeta_1 \\ 
\vdots \\
\zeta_9
\end{bmatrix} = \Phi^T \vec{\zeta}.
\end{align}
where
\begin{equation}
\Phi^T  = \begin{bmatrix}
\xi_R & \xi_R^2 & \xi_R^3 & \xi_L & \xi_L^2 & \xi_L^3 & sin(\theta) &  cos(\theta) & |\dot{\theta}|
\end{bmatrix}^T
\end{equation}
 is the RLS regressor matrix,  and $\vec{\zeta} \in {\rm I\!R}^{9}$ is known as the parameter vector.  A recursive least squares solution for $\vec{\zeta}$ can be found in many works including \cite{plackett1950RLS}.  The necessary and sufficient conditions for global asymptotic stability of RLS is described in \cite{Bruce2021stabilityRLS}.

 \section{IMPLEMENTATION DETAILS}
These three approaches were implemented onboard Heron vehicles which were operating on the Charles River throughout the summer and fall of 2022.  The front-seat/back-seat paradigm was used for autonomous control of each vehicle, where the backseat autonomy platform was MOOS-IvP \cite{BenjaminMOOS}, which uses the MOOS publish-subscribe middleware and behavior optimization via IvP.  The autonomy payload computer was a Rasberry Pi 4.  The forward velocity was measured using GPS when the vehicle speed was greater than $0.2$ m/s, otherwise the velocity estimate was provided by the manufacturer's extended Kalman filter (EKF) estimate of the state. The resolution for the forward velocity data was set at $0.05$ m/s, and this effectively established a lower bound for online prediction error. 

A separate MOOS app, pDynamLearning, was developed to implement these algorithms.  The app pDynamLearning was designed to handle all the communications with the MOOSDB, perform error-checking, load parameters from the last run on startup, and periodically save parameters to the local drive.  The app also provided feedback to the operator by reporting in real-time if the RNN had the contraction property - by evaluating the inequality (\ref{eq:simpleConst}), and displaying the equilibrium point. 
An block diagram of the implementation is shown in the inset of Figure \ref{fig:HeronRobots}, and a summary of number of parameters is provided in Table \ref{tab:params_complex}.

\begin{table}[ht]
	\caption{Implementation parameters and complexity }
	\label{tab:params_complex}
		\begin{tabular}{clll} 
             
		              &      & Forward & Parameter \\ 
            Method   & Implemented  & Estimate & Update \\
                     &  Size              &   Complexity  & Complexity \\
                                 
            \hline
			\rule{0pt}{3ex}  AID &  $p_{AID}=9$   &  $O(p_{AID})$ & $O(p_{AID})$     \\
               \vspace{1mm}
                &    parameters &  &  \\

			RNN &  $n=20$, $m=4$;  & $O(n \times (m+1))$  & $O(n \times (m+1))$ \\
            \vspace{1mm}
                &  $140$ weights &  & \\
   
			RLS &  $p_{RLS}=9$  & $O(p_{RLS})$ &  $O(p_{RLS}^2)$  \\ 	
             &    parameters &  &  \\        
		\end{tabular}
\end{table}

The app pDynamLearning subscribed to the forward speed, $v_k$, and heading $\theta_h$, and also subscribes to the left and right thruster commands, $\xi_L$ and $\xi_R$.  The timestamps associated with the messages which contain the three input variables, $\theta_h$, $\xi_L$, $\xi_R$ were recorded.  To be considered a complete set of inputs, every message must arrive within a set window of time, which was set to $0.2$ seconds.  If this staleness condition was met, the inputs were used to estimate the velocity of the next time step $v_{k+1}$, and for the AID and RNN algorithms, this calculation used the previous estimate from the AID and RNN respectively.  In this way the AID and RNN velocity estimates were programmed as dynamical systems, which clearly illustrates the need for these two methods to be stable.  An average of the three estimates was calculated, and all estimates were published. 

If the measured ``true'' velocity was also received within the set window of time, in addition to all the input messages, then the process of updating the weights was performed in each of the three methods.  The state variable for the RNN was scaled as described in Section \ref{sec:cont_USV_motion}.  Additionally, the app handled edge cases associated with startup or the intermittent periods of time where the measured velocity was not available. The computational complexity associated with the parameter update (AID), back propagation (RNN), and weight update (RLS) is shown in Table \ref{tab:params_complex}.

To enable continuous estimation over months of deployments, the app was programmed to save and load the current model parameters and weights. The app was programmed to save the most recent parameter estimates and weights to the local back-seat computer every five minutes of run time.  On start-up the app loaded the most recent parameter set found on the local disk. The first time the app was launched in 2022 it loaded parameters identified via offline training using 29 hours of data collected from the 2021 season, but every subsequent launch it loaded parameters that were calculated and saved during the previous mission.  In this way the system identification continued over the course of months of deployments.  Finally, this data was periodically offloaded for further analysis.

\section{EXPERIMENTAL RESULTS}
Throughout the summer and fall of 2022, the performance of each method was evaluated using eight Heron ASVs.  The eight ASVs were involved in demonstrations of missions related to convoying, patrolling, surveying that took place on the Charles River near the MIT sailing pavilion in Cambridge, MA, and at Lake Popolopen in West Point, NY.  Typically, the pDynamLearning app was launched as an ancillary MOOS app on each vehicle in the back-seat MOOS community, and would run in the background while the mission was executed.  During the season, we collected 30 hours of run data, including many operations that were aborted mid-mission, paused, or otherwise disrupted.  The intention was to evaluate these methods against actual real-world operational conditions, and not only missions that are carefully curated and well executed. 

This desire to evaluate performance on any and all missions meant that the online learning algorithms encountered many outliers in the incoming data, and almost certainly noise that was not zero-mean Gaussian.  As a consequence, there were outliers in the error between the measured velocity and the estimated velocity from each method.

\subsection{Performance of online prediction during 2022 season}
\begin{figure}[h]
  \centering
  \includegraphics[width=1\columnwidth]{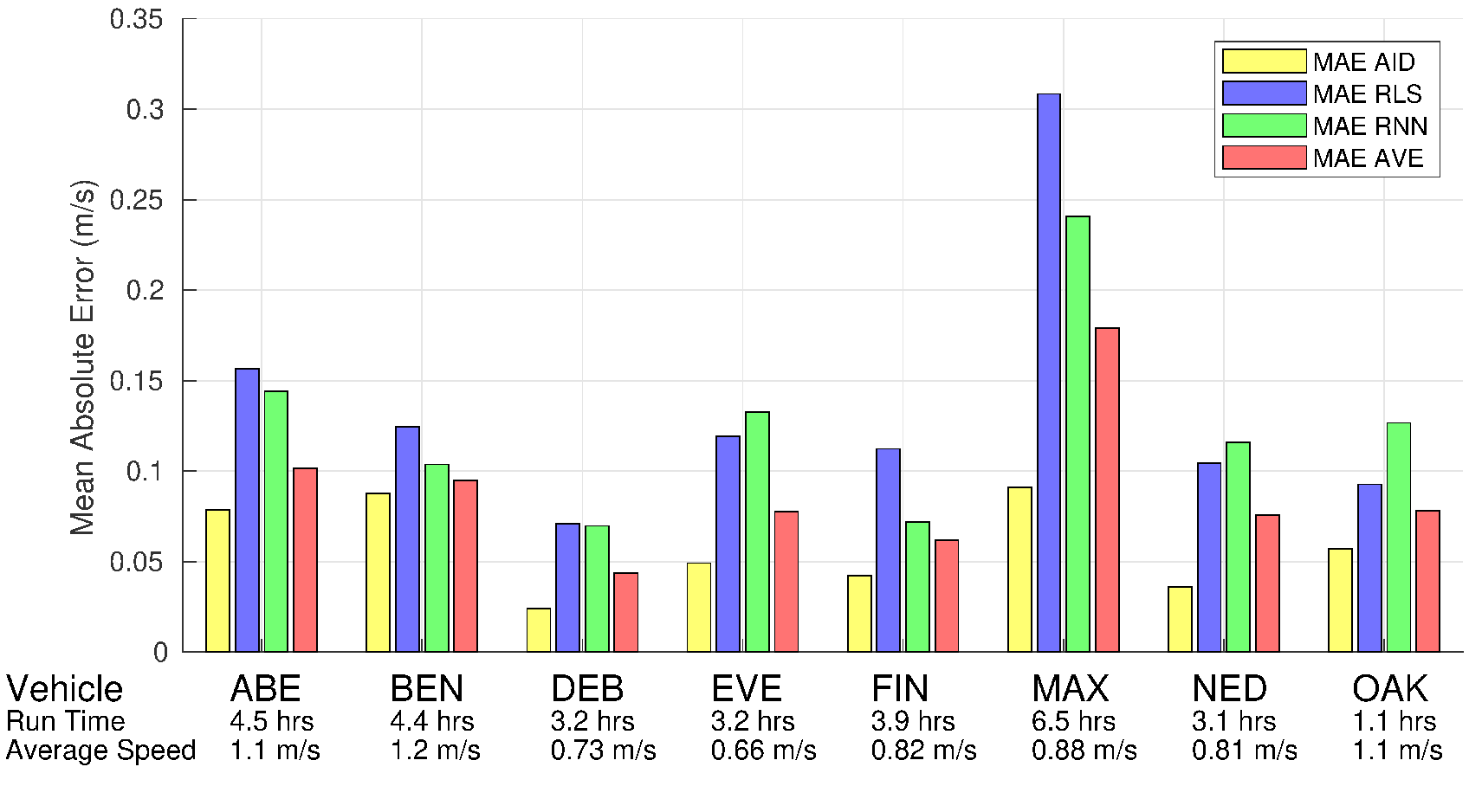}
  \caption{Online prediction error for each of eight vehicles using all three methods (AID, RLS, RNN) and the basic ensemble average (AVE).  Total run time and average speed are listed below vehicle name.}
  \label{fig:EqnError}
  \vspace{-3mm}
\end{figure}

The performance of each algorithm in online estimation is shown Figure \ref{fig:EqnError},   In the plot we show the real-time prediction error, i.e. the error in predicting the forward velocity at the next time step $v_{k+1}$ given the current state and inputs.  The error metric is the mean absolute error (MAE).

\begin{figure*}
  \includegraphics[width=\textwidth]{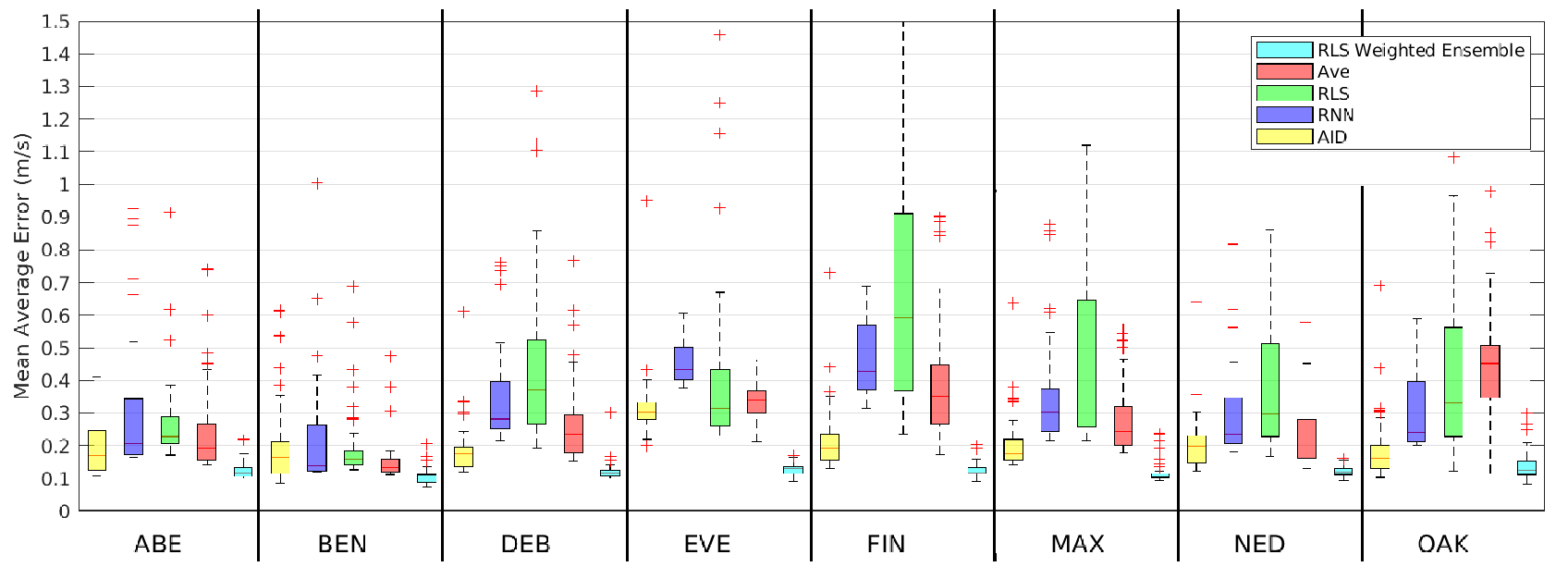}
  \caption{Cross validation results for each vehicle and each of the four online methods and the offline RLS weighted ensemble method.  For each vehicle the distribution is calculated in a cross validation using all saved parameter sets and all run data. Except for one vehicle (Ben), the AID method had lowest MAE of all online methods, but the RLS weighted ensemble had lower error in offline processing.}
  \label{fig:CrossVallAll8}
  \vspace{-4mm}
\end{figure*}

On all eight vehicles the AID method had the lowest prediction error, and the simple average of the outputs from the three methods did not result in less MAE than just using the AID method alone. 

\subsection{Offline data evaluation}
Although all parameter or weight updates were performed online, and the algorithms provided a real-time prediction of the surge velocity at the next time step $v_{k+1}$, we evaluated the stability and the accuracy of each method using offline cross validation.   Cross validation is a process where the parameter and weight estimates recorded by pDynamLearning during one mission are used in a forward simulation with inputs from other missions.  The process is akin to the practice of using different training and testing datasets.  Here we used all of the saved sets of parameters and weights, including those that were saved intermittently throughout the mission. The total number of sets saved by each vehicle varied with a range from 43 to 82.  As a result, we can evaluate performance during each mission and also across all missions. 

The errors in cross validation can be seen in Figure \ref{fig:CrossVallAll8}.  The AID method performed the best on average for seven of eight vehicles and had the most consistent performance - smallest range between the minimum and maximum error - on six of eight vehicles.  There were many outliers in the validation results when the RNN and RLS methods were used, and the AVE did not perform better than just the AID method alone.   However, we were interested to see if a weighted ensemble would perform better.

\subsubsection{Weighted ensemble}
In the offline setting we experimented with fusing together the three estimates of forward velocity: $v_{AID_{k+1}}$, $v_{RNN_{k+1}}$, $v_{RLS_{k+1}}$ at each timestep where the weighted ensemble velocity
\begin{equation}
    v_{WE_{k+1}} =  \bar{c}_1 v_{AID_{k+1}} + \bar{c}_2 v_{RNN_{k+1}} + \bar{c}_3 v_{RLS_{k+1}}.  
\end{equation}
The weights $\bar{c}_1, \bar{c}_2, \bar{c}_3$ are estimated using another RLS optimization. Even though this weighting was done offline, the RLS estimation process was completed at each time-step in the same way it would be completed online.   We found this method works particularly well at fusing the outputs of the three models, and the quantitative results in cross validation can be seen in Figure \ref{fig:CrossVallAll8} with the label ``RLS Weighted Ensemble''.   We plan to deploy this approach in the future.

\subsubsection{Cross validation between vehicles}
\begin{figure}[h]
  \centering
  \includegraphics[width=1\columnwidth]{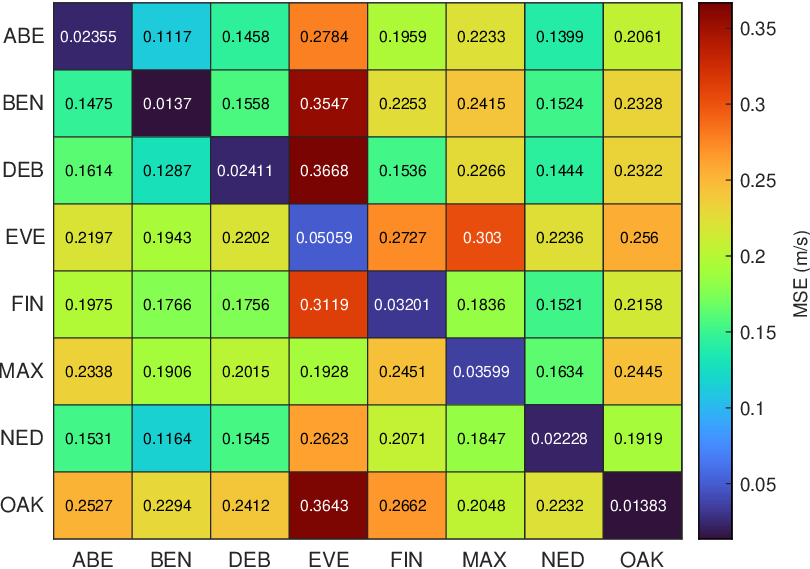}
  \caption{Cross validation error between vehicles, i.e. the best model identified for one vehicle (rows) is cross validated with run data from all other vehicles (columns).  The entries on the diagonal are self validation error with all run data from own vehicle.  The error metric is the mean squared error (MSE) of the average (AVE) of the three system identification methods. }
  \label{fig:CrossValDiffVeh8}
  \vspace{-5mm}
\end{figure}

Additionally, we can cross validate between vehicles to show the learned models are uniquely optimized for each vehicle.  In this analysis we take the best set of parameters for each vehicle and compare the performance of that set on each of the other seven vehicles.  The result is shown in Figure \ref{fig:CrossValDiffVeh8}, where we use the mean squared error (MSE) instead of MAE to draw better contrast in performance.  The results in Figure \ref{fig:CrossValDiffVeh8} are encouraging because we initialized the parameters to be the same on every vehicle at the start of the 2022 season.  As the season progressed the online system identification methods optimized the parameters to reflect the unique dynamics of each vehicle.  This result confirms our observations about the individual performance of some vehicles during the season; for example some vehicles have a noticeable thruster bias, and others do not. 

\subsection{Discussion}
Although all methods demonstrated stable performance during the 2022 season, the AID method consistently had the best performance with respect to equation error and was the best in seven of eight vehicles with respect to cross validation error. We offer some opinions about why this was the case. One reason the AID approach was more stable is likely due to the choice of adaptation gains, and perhaps better robustness to non-Gaussian noise than the RLS or RNN methods.  In contrast, the RLS method minimizes the sum of the square of the residuals, which can be significantly skewed by outliers.  Certainty, for the RLS approach the disruption of outliers on performance is a function of the ``forgetting factor'', and this hyperparameter could be further optimized in these experiments.  Although in an offline setting it is possible to design and tune an RNN to match the training data with minimal error, it was a challenge to find parameters such that the RNN performance would generalize in the online setting as it encounters new training data in real-time.   

All hyper-parameters were selected using offline training with approximately 29 hours of run data from the previous 2021 operating season. The values were selected using a coarse grid search followed by a process of refinement that used both the understanding of the theoretical effect of each hyperparameter and empirical evaluation.   

Finally, we chose to implement a shallow RNN architecture over other popular architectures such as a deep neural network (DNN) or LSTM primarily because of the properties of the dynamical system we were estimating.  The common plant model of USV motion (\ref{eq:simple1dof}) described the evolution of $v(t)$ as a first order linear differential equation. Simply put, in discrete form the state at the next time step $v_{k+1}$ only depended on the value of the state at the previous time step $v_{k}$ and the input at the next time step $\xi_{L_{k+1}}$, $\xi_{R_{k+1}}$.  For this reason it was appropriate to use a shallow RNN architecture as a direct comparison to the AID and RLS methods.  Although we were optimistic the RNN would perform best, since it could theoretically better model the effects of unknown non-linear dynamics, it did not.   However we acknowledge that a different choice of RNN hyper parameters could result in better performance, and that is the goal of future work.  

\section{CONCLUSIONS}
This paper presented an experimental comparison of three popular methods for online system identification for USVs with over 30 hours of operational data during the 2022 season.  In general, the AID approach had more stable performance.  Averaging the velocity estimates from all three methods did not always improve the accuracy, but we  showed in an offline setting it can be beneficial to fuse together the outputs from all three methods using recursive linear regression.  However, it was difficult to make a fair comparison because the performance of each method is so dependent on the magnitude of the tuning parameters: adaptation gains, ``forgetting factor'', learning rate and ADAM parameters.






\section*{ACKNOWLEDGMENTS}
We thank Professor Jean-Jacques Slotine for a helpful discussion regarding contraction analysis.  We also thank Raymond Turrisi and Kevin Becker for supporting deployments of the Heron fleet.  The image in Figure \ref{fig:HeronRobots} was taken by Raymond Turrisi. 

\bibliographystyle{IEEEtran}
\bibliography{IEEEabrv, refs_updated}

\begin{thebibliography}{10}
\providecommand{\url}[1]{#1}
\csname url@rmstyle\endcsname
\providecommand{\newblock}{\relax}
\providecommand{\bibinfo}[2]{#2}
\providecommand\BIBentrySTDinterwordspacing{\spaceskip=0pt\relax}
\providecommand\BIBentryALTinterwordstretchfactor{4}
\providecommand\BIBentryALTinterwordspacing{\spaceskip=\fontdimen2\font plus
\BIBentryALTinterwordstretchfactor\fontdimen3\font minus
  \fontdimen4\font\relax}
\providecommand\BIBforeignlanguage[2]{{%
\expandafter\ifx\csname l@#1\endcsname\relax
\typeout{** WARNING: IEEEtran.bst: No hyphenation pattern has been}%
\typeout{** loaded for the language `#1'. Using the pattern for}%
\typeout{** the default language instead.}%
\else
\language=\csname l@#1\endcsname
\fi
#2}}

\bibitem{RandeniP2018NonlinearDynamics}
S.~A.~T. Randeni~P., A.~L. Forrest, R.~Cossu, Z.~Q. Leong, D.~Ranmuthugala, and
  V.~Schmidt, ``Parameter identification of a nonlinear model: replicating the
  motion response of an autonomous underwater vehicle for dynamic
  environments,'' \emph{Nonlinear Dynamics}, vol.~91, no.~2, pp. 1229--1247,
  Jan 2018.

\bibitem{birnbaum2014UAS}
Z.~Birnbaum, A.~Dolgikh, V.~Skormin, E.~O'Brien, and D.~Muller, ``Unmanned
  aerial vehicle security using recursive parameter estimation,'' in \emph{2014
  International Conference on Unmanned Aircraft Systems (ICUAS)}, 2014, pp.
  692--702.

\bibitem{ijoe2014martin}
S.~C. Martin and L.~L. Whitcomb, ``{Experimental Identification of
  Six-Degree-of-Freedom Coupled Dynamic Plant Models for Underwater Robot
  Vehicles},'' \emph{IEEE Journal of Oceanic Engineering}, vol.~39, no.~4, pp.
  662--671, 10 2014.

\bibitem{2007hegrenaes}
O.~Hegren{\oe}s, O.~Hallingstad, and B.~Jalving, ``{Comparison of Mathematical
  Models for the HUGIN 4500 AUV Based on Experimental Data},'' in \emph{2007
  Symposium on Underwater Technology and Workshop on Scientific Use of
  Submarine Cables and Related Technologies}, 4 2007, pp. 558--567.

\bibitem{graver2003underwater}
J.~G. Graver, R.~Bachmayer, N.~E. Leonard, and D.~M. Fratantoni, ``Underwater
  glider model parameter identification,'' in \emph{Proc. 13th Int. Symp. on
  Unmanned Untethered Submersible Technology (UUST)}, vol.~1, 2003, pp. 12--13.

\bibitem{cst2003smallwood}
D.~A. Smallwood and L.~L. Whitcomb, ``{Adaptive Identification of Dynamically
  Positioned Underwater Robotic Vehicles},'' \emph{IEEE Transactions on Control
  Systems Technology}, vol.~11, no.~4, pp. 505--515, July 2003.

\bibitem{paine2018Oceans}
T.~M. Paine and L.~L. Whitcomb, ``Adaptive parameter identification of
  underactuated unmanned underwater vehicles: A preliminary simulation study,''
  in \emph{OCEANS 2018 MTS/IEEE Charleston}.\hskip 1em plus 0.5em minus
  0.4em\relax IEEE, 10 2018, pp. 1--6.

\bibitem{mcfarlandJAC2021}
C.~J. McFarland and L.~L. Whitcomb, ``Stable adaptive identification of
  fully-coupled second-order 6 degree-of-freedom nonlinear plant models for
  underwater vehicles: Theory and experimental evaluation,''
  \emph{International Journal of Adaptive Control and Signal Processing}, 2021.

\bibitem{harrismaopaine2022IJRR}
A.~M. Mao*, Z.~J. Harris*, T.~M. Paine*, and L.~L. Whitcomb, ``Stable
  null-space adaptive parameter identification of 6 degree-of-freedom decoupled
  plant and actuator models for underactuated vehicles: Theory and experimental
  evaluation,'' \emph{Accepted by the International Journal of Robotics
  Research}.

\bibitem{Wehbe2017icra}
B.~Wehbe, M.~Hildebrandt, and F.~Kirchner, ``Experimental evaluation of various
  machine learning regression methods for model identification of autonomous
  underwater vehicles,'' in \emph{2017 IEEE International Conference on
  Robotics and Automation (ICRA)}, May 2017, pp. 4885--4890.

\bibitem{Wehbe2019icra}
------, ``A framework for on-line learning of underwater vehicles dynamic
  models,'' in \emph{2019 International Conference on Robotics and Automation
  (ICRA)}, 2019, pp. 7969--7975.

\bibitem{sonnenburg2013JFR}
C.~R. Sonnenburg and C.~A. Woolsey, ``Modeling, identification, and control of
  an unmanned surface vehicle,'' \emph{Journal of Field Robotics}, vol.~30,
  no.~3, pp. 371--398, 2013.

\bibitem{yuh1999ICRA}
J.~Yuh, J.~Nie, and C.~Lee, ``Experimental study on adaptive control of
  underwater robots,'' in \emph{Proceedings 1999 IEEE International Conference
  on Robotics and Automation (Cat. No.99CH36288C)}, vol.~1, 1999, pp. 393--398
  vol.1.

\bibitem{woo2018AOR}
J.~Woo, J.~Park, C.~Yu, and N.~Kim, ``Dynamic model identification of unmanned
  surface vehicles using deep learning network,'' \emph{Applied Ocean
  Research}, vol.~78, pp. 123--133, 2018.

\bibitem{miller2018stable}
J.~Miller and M.~Hardt, ``Stable recurrent models,'' in \emph{International
  Conference on Learning Representations}, 2018.

\bibitem{lohmiller1998contraction}
W.~Lohmiller and J.-J.~E. Slotine, ``On contraction analysis for non-linear
  systems,'' \emph{Automatica}, vol.~34, no.~6, pp. 683--696, 1998.

\bibitem{fossen_1994}
T.~I. Fossen, \emph{{G}uidance and {C}ontrol of {O}cean {V}ehicles}.\hskip 1em
  plus 0.5em minus 0.4em\relax New York, NY: John Wiley and Sons, 1994.

\bibitem{TSUKAMOTO2021135}
H.~Tsukamoto, S.-J. Chung, and J.-J.~E. Slotine, ``Contraction theory for
  nonlinear stability analysis and learning-based control: A tutorial
  overview,'' \emph{Annual Reviews in Control}, vol.~52, pp. 135--169, 2021.

\bibitem{martinthesis}
S.~C. Martin, ``{Advances in Six-Degree-of-Freedom Dynamics and Control of
  Underwater Vehicles},'' Ph.D. dissertation, {The Johns Hopkins University},
  2008.

\bibitem{manchester2021CiRNNTutorial}
I.~R. Manchester, M.~Revay, and R.~Wang, ``Contraction-based methods for stable
  identification and robust machine learning: a tutorial,'' in \emph{2021 60th
  IEEE Conference on Decision and Control (CDC)}, 2021, pp. 2955--2962.

\bibitem{Revay2020ContractingRNN}
M.~Revay and I.~Manchester, ``Contracting implicit recurrent neural networks:
  Stable models with improved trainability,'' in \emph{Proceedings of the 2nd
  Conference on Learning for Dynamics and Control}, ser. Proceedings of Machine
  Learning Research, A.~M. Bayen, A.~Jadbabaie, G.~Pappas, P.~A. Parrilo,
  B.~Recht, C.~Tomlin, and M.~Zeilinger, Eds., vol. 120.\hskip 1em plus 0.5em
  minus 0.4em\relax PMLR, 10--11 Jun 2020, pp. 393--403.

\bibitem{jin1999stable}
L.~Jin and M.~M. Gupta, ``Stable dynamic backpropagation learning in recurrent
  neural networks,'' \emph{IEEE Transactions on Neural Networks}, vol.~10,
  no.~6, pp. 1321--1334, 1999.

\bibitem{horn2012matrix}
R.~Horn and C.~Johnson, \emph{Matrix Analysis}.\hskip 1em plus 0.5em minus
  0.4em\relax Cambridge University Press, 2012.

\bibitem{galor2007discrete}
O.~Galor, \emph{Discrete dynamical systems}.\hskip 1em plus 0.5em minus
  0.4em\relax Springer Science \& Business Media, 2007.

\bibitem{ClearpathHeronRepo}
``Heron,'' \url{https://github.com/heron/heron}, Clearpath Robotics, 2022.

\bibitem{plackett1950RLS}
R.~L. Plackett, ``Some theorems in least squares,'' \emph{Biometrika}, vol.~37,
  no. 1/2, pp. 149--157, 1950.

\bibitem{Bruce2021stabilityRLS}
A.~L. Bruce, A.~Goel, and D.~S. Bernstein, ``Necessary and sufficient regressor
  conditions for the global asymptotic stability of recursive least squares,''
  \emph{Systems \& Control Letters}, vol. 157, p. 105005, 2021.

\bibitem{BenjaminMOOS}
M.~R. Benjamin, H.~Schmidt, P.~M. Newman, and J.~J. Leonard, ``Nested autonomy
  for unmanned marine vehicles with moos-ivp,'' \emph{Journal of Field
  Robotics}, vol.~27, no.~6, pp. 834--875, November/December 2010.

\end{thebibliography}

\end{document}